\documentclass[USenglish]{article}	
\usepackage[utf8]{inputenc}				%(only for the pdftex engine)
\usepackage[big,online]{dgruyter}	%values: small,big | online,print,work
\usepackage{lmodern} 
\usepackage{microtype}
\usepackage[round,authoryear,sort&compress]{natbib}
\usepackage{graphicx}
\usepackage{subcaption}

\theoremstyle{dgthm}

\theoremstyle{dgdef}

\begin{document}

%%%--------------------------------------------%%%
	% \articletype{Research Article}
	% \received{Month	DD, YYYY}
	% \revised{Month	DD, YYYY}
 %  \accepted{Month	DD, YYYY}
  % \journalname{Journal~of~Sports~Analytics}
  % \journalyear{YYYY}
  % \journalvolume{XX}
  % \journalissue{X}
  \startpage{1}
  \aop
  % \DOI{10.1515/sample-YYYY-XXXX}
%%%--------------------------------------------%%%

\title{Assessing win strength in MLB win prediction models}
\runningtitle{Assessing win strength in MLB win prediction models}
%\subtitle{Insert subtitle if needed}

\author[1]{Morgan Allen}
%\ use * to mark the author as the corresponding author
\author*[2]{Dr. Paul Savala}
\affil[1]{\protect\raggedright 
St. Edward's University, e-mail: morgan.allen@outlook.com}
\affil[2]{\protect\raggedright 
St. Edward's University, e-mail: psavala@stedwards.edu}
	
%\communicated{...}
%\dedication{...}
	
\abstract{In Major League Baseball, strategy and planning are major factors in determining the outcome of a game. Previous studies have aided this by building machine learning models for predicting the winning team of any given game. We extend this work by training a comprehensive set of machine learning models using a common dataset. In addition, we relate the win probabilities produced by these models to win strength as measured by score differential. In doing so we show that the most common machine learning models do indeed demonstrate a relationship between predicted win probability and the strength of the win. Finally, we analyze the results of using predicted win probabilities as a decision making mechanism on run-line betting. We demonstrate positive returns when utilizing appropriate betting strategies, and show that naive use of machine learning models for betting lead to significant loses.}


\maketitle

\section{Introduction} 
Baseball is a game dominated by statistics and measurements that was initiated by Billy Beane as the general manager of the 2002 Oakland Athletics. However, Baseball is no longer just about analytics but is a complex, inter-working system of models. Teams use complex algorithms, machine learning, and deep learning models across every aspect of baseball interaction. Examples include predicting player development and output [\cite{elitzur2020data}], using computer vision to track pitches and faults in mechanics [\cite{healey2017new}], and tracking each swing with Statcast tools [\cite{lage2016statcast}]. These models extend beyond the playing surface into the front office for modeling ticket sales and many other uses [\cite{davenport2014businesses}]. With new sensors and tracking technologies, a plethora of data is waiting to be sifted through and used to gain a competitive edge.

In our work we first train the most common machine and deep learning models on individual player and team statistics to predict team wins and loses. While several previous studies have done the same, in many cases they train only a small subset of common models, train models on data which is too small to generalize, or make methodological mistakes which make their conclusions suspect. In this work we remedy this by collecting a large number of player and team statistics. We also ensure no overlap in our training and testing sets by splitting by season.

Beyond model development, our primary interest is in understanding the extent to which models can infer the strength of a win. For example, suppose teams A and B are playing one another. Furthermore, suppose a model predicts that team A will win with a predicted probability of 90\%. In that case, one may intuitively expect team A to win by a sizeable margin. On the other hand, if team A only has a predicted win probability of 51\%, one may expect the game to be close in final score differential. However, no work to date has investigated the extent to which this is true. Do the predicted win probabilities produced by models capture this relationship between win probability and a measure of winning such as score differential? Said another way, are the predicted win probabilities helpful in understanding the strength of a win?

Throughout this paper, our models are classification models which predict whether the home team will win or lose. We also capture the predicted probability of a home team win produced by each model. We show there is an important and predictable relationship between predicted win probability and the strength of a win.

\section{Background} 

\subsection{Literature review}	
A number of studies have examined the ability of machine learning models to predict wins in Major League Baseball. A summary of these results is shown in Table \ref{tab:LitReviewModels}.

In unpublished work, \cite{jia2013predicting} trained logistic regression (LogR), support vector machine (SVM) and gradient boosting (GB) models on the 2007 to 2011 MLB seasons. They achieve accuracy of approximately 56\% through three different models, namely a support vector machine and two gradient boosting models. A notable outcome of this study is that the authors realized that more recent statistics from the current season were more impactful than lifetime statistics for each player. 

\cite{soto2016predicting} trained artificial neural network (ANN), SVM, decision tree (DT), and a K-nearest neighbors (KNN) models to predict game wins. They used data from the 2005 to 2014 seasons and achieved a peak accuracy of 58.92\% using a form of an SVM. However, they used 10-fold cross validation, which randomly keeps one-tenth of the data aside for testing purposes. One drawback to this approach is that the seasons from the testing data almost certainly overlap with seasons from the training data. This approach has theoretical problems with respect to temporal dependence [\cite{bergmeir2012use}].

\cite{elfrink2018predicting} trained random forest (RF), GB, linear regression (LR), and LogR models for the same purpose. They achieved accuracy of 55.52\% using a gradient boosting model. The authors trained their models on the 1930 to 2015 seasons and tested on the 2016 season. Their work suggests ensembling many models to achieve a rise in accuracy due to the models predicting in disparate ways.

\cite{huang2021use} considered the effect of moving away from classical machine learning models and focusing on various neural network architectures. They fit SVM and several neural networks to predict outcomes, and achieved at least 90\% accuracy for each model. A notable shortcoming however is that they only used the 2019 season, and used it for both training and testing using cross-validation. This perhaps explains the very large difference in model performance relative to past studies, even for the same models and similar data to past studies. They used cross-validation within the 2019 season for training and testing, likely leading to significant temporal dependence relationships which may not have been taken into account.

While we focus here on models predicting wins, some authors have considered the prediction of runs. One such work is \cite{beneventano2012predicting} who trained linear regression models on the 2001 to 2010 MLB seasons to predict runs. 

\begin{table} [h]
\caption{Comparison of win prediction models}
\begin{tabular}{lcccc}
Paper 	& Models compared 	        & Seasons trained on    & Seasons tested on  & Peak accuracy (model)	    \\ \midrule
\cite{jia2013predicting} & SVM, GB & 2007-2011 & 2012 & 59.6\% (SVM) \\
\cite{soto2016predicting} & ANN, SVM, DT, KNN & 2005-2014 & 2005-2014 (10-fold CV) & 58.92\% (SVM) \\
\cite{elfrink2018predicting} & RF, GB, LR, LogR & 1930-2015 & 2016 & 55.52\% (GB)	\\
\cite{huang2021use}	& SVM, ANN & 2019 & 2019 (5-fold CV) & 94.18\% (ANN) \\
\end{tabular}
\label{tab:LitReviewModels}
\end{table}

\subsection{Data description}
For this study, the data spans the 2001 to 2019 seasons and does not include the playoffs. These years were selected to keep the models as relevant as possible to modern baseball. The data for this research was sourced from Baseball Reference [\cite{baseball_reference}], FanGraphs [\cite{fangraphs}], Lahman's Baseball Dataset [\cite{lahman}], and Retrosheet [\cite{retrosheet}]. All of these sources are well-known within the community and used by papers in the literature review section. There are a total of 46,159 games in our dataset. 

Table \ref{tab:Variables} shows all of the features that were used as inputs to the models. Each variable includes the information shown in the ``Variable'' column, and some variables contain variations on these variables, such as the percent different between the two teams playing (PctDiff), the raw difference between the teams (Diff), and the previous years value (Offset 1 Yr). Abbreviations for these variations are listed at the bottom of the table. The hitting variables contain team data for both the current season and some inputs for the previous season. There pitching variables represent the bullpen and overall pitching staff and starting pitcher variables for the given season. The other input variables are many measures of win differential and probabilities as well as popular metrics like average attendance, ELO ratings by FiveThirtyEight [\cite{fte_github_mlb_elo}], and Pythagorean Expectation. In Table \ref{tab:CompInputVariables} we compare our variables to those from each study presented in the literature review. Our work uses 83 variables compared to the next closest study only using 30 variables.

\begin{table}[h]
\caption{Variables used in fitting models}
\begin{tabular}{llcccc}
Type               & Variable                                     & Abbreviation & Pct Diff & Diff & Offset 1 Yr \\ \midrule
Hitting Variables  & Team OPS                                     & OPS          & \checkmark        & \checkmark    &                 \\ \cline{2-6} 
& Team AVG                                     & AVG          & \checkmark        & \checkmark    &                  \\ \cline{2-6} 
                   & Team SLG                                     & SLG          & \checkmark        & \checkmark    &                  \\ \cline{2-6} 
                   & Team OBP                                     & OBP          & \checkmark        & \checkmark    &                  \\ \cline{2-6} 
                   & Team Runs                                    & R            & \checkmark        &     &                  \\ \cline{2-6} 
                   & Team Fielding Percentage                     & FP           & \checkmark        &     & \checkmark                 \\ \cline{2-6} 
                   & Run Differential                             & RD           &         &     &                 \\ \cline{2-6} 
                   & ISO                                          & ISO          &         &     &              \checkmark    \\ \cline{2-6} 
                   & Total Runs                                   & TotalR       &         &     & \checkmark                 \\ \hline
Pitching Variables & Team ERA                                     & ERA          & \checkmark        &     & \checkmark                \\ \cline{2-6} 
                   & Team WHIP                                    & WHIP         & \checkmark        &     & \checkmark                 \\ \cline{2-6} 
                   & Team Runs Allowed                            & RA           & \checkmark        &     & \checkmark                 \\ \cline{2-6} 
                   & Starting Pitcher Innings Pitched             & SP-IP        & \checkmark        &     &                 \\ \cline{2-6} 
                   & Starting Pitcher WPA                         & SP-WPA       & \checkmark        &     &                  \\ \cline{2-6} 
                   & Starting Pitcher ERA                         & SP-ERA       & \checkmark        &     &                  \\ \cline{2-6} 
                   & Starting Pitcher WHIP                        & SP-WHIP      & \checkmark        &     &                  \\ \cline{2-6} 
                   & Starting Pitcher \# Games Thrown This Season & SP-NumG      &         &     &                 \\ \hline
Other Variables    & Team Bayes Win Probability                  & BayesWP      & \checkmark        &     & \checkmark                 \\ \cline{2-6} 
                   & Team W-L                                     & W-L          &         &     & \checkmark                 \\ \cline{2-6} 
                   & Team Rank                                    & Rank         &         &     & \checkmark                \\ \cline{2-6} 
                   & Team Pythagorean Expectation                 & TeamPythag   &         &     &                \\ \cline{2-6} 
                   & Average Attendance                           & AvgAttend    &         &     & \checkmark                 \\ \cline{2-6} 
                   & Team Win Differential                        & WD           &         &     &                  \\ \cline{2-6} 
                   & Team Win Percentage                          & WP           &         &     &                 \\ \cline{2-6} 
                   & Team ELO (FiveThirtyEight)                   & ELO          &         &     &                  \\ \cline{2-6} 
                   & Team Rest Days                               & RD           &         &     &                 \\ \cline{2-6} 
                   & Team Previous Game WL                        & PG-WL        &         &     &                  \\ \cline{2-6} 
                   & Log5                                         & Log5         &         &     &              \\ \cline{2-6} 
                   & Year                                         & Y            &         &     &               \\ \cline{2-6} 
                   & Month                                        & M            &         &     &               \\ \hline
Abbreviations & & * & *pctDiff & *Diff & *-1  \\ 
\end{tabular}
\label{tab:Variables}
\end{table}

\begingroup
\renewcommand{\arraystretch}{1.5}
\begin{table} [h]
\caption{Comparison of Input variables}
\begin{tabular}{lp{10cm}}
Paper	 & Input variables 	                     	    \\ \midrule
\cite{jia2013predicting} & AVG, RBI, OBP, ERA, H, E, and Win\% for each team \\
\hline
\cite{soto2016predicting} & isHomeClub, Log5, PE, WP, RC, PrevWL, BABIP, FP, PitchERA, OBP, SLG, VisitorLeague, HomeVersusVisitor, Stolen \\
\hline
\cite{elfrink2018predicting} & AB, AVG, OBP, SLG, OPS, AVG/RISP, WHIP, RA	\\
\hline
\cite{huang2021use}	& 
AB, H, BB, PA, AVG, OBP, SLG, OPS, Pit, Str, PO, IP, H, BB, HR, ERA, BF, Pit, Str, Ctct, FB, Gsc, IR, IS, H/A
 \\
\hline
Allen and Savala (2022)	& 
OPSpctDiff, SLGpctDiff, OBPpctDiff, AVGpctDiff, RpctDiff, FPpctDiff, OPSDiff, SLGDiff, OBPDiff, AVGDiff, OPS, SLG, OBP, AVG, R, RD, ISO, FP-1, R-1, ERApctDiff, WHIPpctDiff, RApctDiff, SP-IPpctDiff, SP-WPApctDiff, SP-ERApctDiff, SP-WHIPpctDiff, WHIP-1, ERA-1, RA-1, RA, SP-ERA, SP-WHIP, SP-WPA, SP-IP, SP-NumG , BayespctDiff, W-LpctDiff, RankpctDiff, PythagpctDiff, RDpctDiff, Rank-1, W-L-1, Attend-1, Bayes, WD, Pythag, WP, ELO, Rest, PrevWL, Log5, Y, M
 \\
\end{tabular}
\label{tab:CompInputVariables}
\end{table}
\endgroup

\section{Methodology}

\subsection{Models}
In this paper we evaluate the performance of six different families of models: a baseline model which always predicts a home-team win, logistic regression, support vector machines, gradient boosting decision trees, K-nearest neighbors, and artificial neural networks. We also compare our models to a ELO-based model run by FiveThirtyEight. For each family of models we utilized cross-validation grid search when applicable to scan for the best choice of hyperparameters. In each case models were trained on the 2001-2015 seasons and tested on the held-out 2016-2019 seasons. For all seasons, playoffs were excluded as it has been suggested that player performances in the regular season and playoffs can differ quite dramatically [\cite{chu2019empirical}, \cite{conforti2021analysis}].

\subsubsection{Home-team win}
For the sake of comparison we include the null model which always predicts a home team win (HomeWin), regardless of any other information present. It has been shown that being the home team typically leads to a slight advantage [\cite{stefani2008measurement}, \cite{jones2015home}], which is reflected in our data.

\subsubsection{Logistic Regression}
Logistic regression (LogR) is a classical technique which predicts the probability of a given class in a binary prediction problem [\cite{menard:2002}]. In particular, it assumes a linear relationship between the predictor variable(s) and the log-odds (also called ``logit'') of a given value of a binary outcome, such as a win. More concretely, if $x_1, x_2, \ldots, x_n$ denote the features, then logistic regression models choose corresponding parameters $\beta_0, \beta_1, \ldots, \beta_n$ such that $\beta_0 + \beta_1 x_1 + \cdots + \beta_n x_n$ is the least squares regression line for $\ln(p/(1-p))$ where $p$ is the probability of a given outcome of a binary variable. In our case, the class to be predicted is a home team win, and the predictor variables are those described in the data section.

While logistic regression models are generally restricted in their predictive power for data with complex latent relationships [\cite{ranganathan:2017}], they are also interesting because they can be thought of as a single layer artificial neural network [\cite{dreiseitl:2002}]. Therefore, they give a good baseline as to how a neural network may perform. They also are simple to fit with no hyperparameter tuning needed, and their coefficients are directly interpretable [\cite{best:2015}].

\subsubsection{Support Vector Machines}
Support vector machines (SVM) are classification models which separate classes via a hyperplane in the space in which the variables live. By introducing a kernel function the hyperplane can in fact become nonlinear, which allows greater flexibility in separation [\cite{hastie2009elements}]. We use the standard radial basis function kernel. In addition to specifying the kernel one must also specify a regularization parameter. We performed a grid search over values of the regularization parameter between 1 and 10, and chose the optimal parameter value of 1.

\subsubsection{K-Nearest Neighbors}
K-nearest neighbors (KNN) is a clustering-based approach to classification. In its simplest form, KNN starts with a point $x_0$, measures the distance between all other points, finds the $K$ closest points $x_{i_1}, x_{i_2}, \ldots, x_{i_K}$ to $x_0$, and predicts the majority class from $x_{i_1}, x_{i_2}, \ldots, x_{i_K}$ [\cite{laaksonen:1996}]. The predicted probability is obtained by finding the proportion of these $K$ points which correspond to the majority class. So for instance, if $K=10$ and seven of the ten closest points predict a home win, then the predicted probability of a home win is 70\%. In practice, the performance of the model is highly sensitive to the choice of the number of neighbors $K$ and the distance metric. Our choices of these parameters came from a grid search of neighbors between 1 and 300, using the Minkowski distance metric. Optimal predictions occurred with 150 neighbors, and we use that value for all results described below.

\subsubsection{Gradient Boosting}
Gradient boosting is a form of ensemble learning which leverages many weak learners in the form of decision trees to produce a strong learner. In particular, gradient boosted decision trees, which we implement through the software package XGB [\cite{chen:2016}], starts with a single decision tree $T_0$ fitted to the data. A subsequent tree $T_1$ is then introduced in an additive manner by fitting the tree such that $T_0+T_1$ minimizes the loss function. In practice a regularization term is added to prevent overfitting. Gradient boosted decision trees have been shown to be highly effective on a number of tasks, both in sports prediction and in a wide array of other fields. Indeed, several other authors investigating the prediction of sports outcomes have used gradient boosting to great effect [\cite{baboota:2019}, \cite{alfredo:2019}].

\subsubsection{Artificial Neural Network}
Artificial neural networks (ANN) are deep learning models which, in their simplest form, essentially stack multiple logistic regression models on top of one another into a so-called feed-forward neural network. However, their architectures are highly customizable, and therefore can take many different forms. The common theme is taking a linear linear transformation of the inputs and a matrix of coefficients, passing them through a non-linear function such as a sigmoid, and (potentially) repeating this process multiple times. This repetition allows models to learn more and more complex features from the preceding layers [\cite{gurney:2018}]. Neural networks have been shown to be highly effective at a variety of tasks, ranging from tabular learning [\cite{deselits:1992}] to natural language processing [\cite{li:2015}] to computer vision [\cite{khan:2018}]. In addition, neural networks have been widely studied in sports prediction research [\cite{jurgen:2001}, \cite{jurgen:2003}, \cite{baboota:2019}, \cite{alfredo:2019}]. Neural networks are highly sensitive to their architecture. While they have been shown to be universal function approximators [\cite{hornik:1989}], they also have many disparate forms. Therefore a primary concern when constructing a neural network is the architecture. However, when working with tabular data such as ours, feed forward networks have been shown to be highly effective [\cite{borisov:2021}], and thus we use that architecture. A primary concern with neural networks is their lack of interpretability. Much work has been done since their introduction to allow the user to explain the decisions made by neural networks, but this remains a very active area of research [\cite{zhang2021survey}].

\subsubsection{FiveThirtyEight}
While numerous websites and news outlets predict winners, very few assign a probability of winning. The most notable exception is FiveThirtyEight [\cite{fte}], which has its own MLB prediction model. They provide game-level predicted win probabilities, which we can use to compare to our results.

The FiveThirtyEight model (FTE) assigns every team a calculated score representing their skill level, called their ELO. After every game, the winning team gains some number of rating points which depends upon their opponents ELO, and the losing team loses ELO points in a similar manner. While the ELO score forms the basis of the predictions, FiveThirtyEight also adjusts each team's probability of winning by factoring in the home-field advantage, distance traveled to the game, number of days of rest, and the starting pitcher. Home field advantage is worth 24 ELO points, travel is factored in as a linear equation, as is rest. The starting pitcher is given a score based on their strikeouts, outs, walks, hits, runs and home runs. Simulations are then run to determine the probability of winning [\cite{fte:2022}].

\subsection{Measures of win probability}
While our models are trained to predict home team wins and losses, our primary interest is in the relationship between the predicted win probability and the amount by which the team wins or loses, as measured by final score differential.

\subsubsection{Score differential}
The most obvious measure of win strength is score differential, defined as the difference in score between the two teams. For the purposes of this paper we will also compute home team final score minus away team final score. It has been shown that, for a variety of sports including major league baseball, using score differential had improved predictive performance of team rankings compared with only using win-loss data [\cite{barrow2013ranking}]. 

In addition, win differential is used as the input to the statistical measure Pythagorean Expectation, derived by Bill James \cite{james1981baseball}. Pythagorean expectation is defined as 
\begin{equation*}
    \frac{\text{runs scored}^2}{\text{runs scored}^2 + \text{runs allowed}^2}.
\end{equation*}
This formula has since been further analyzed and justified from a statistical viewpoint by \cite{miller2007derivation}. 

\section{Results}
For all models we trained them on the 2001-2015 seasons. We then evaluated the performance of all models on this held out 2016-2019 seasons test set. Recall that models are fit using individual player and team statistics. Therefore, the movement of players between teams after the end of a season does not affect the predictions made, as it is the player and team statistics that form the basis of predictions, and not the teams or players themselves.

\subsection{Model fit} 
In order to assess performance we first considered accuracy. Our data is nearly balanced with the home team winning 53.1\% of the time, thus the null HomeWin model achieves an accuracy of 53.1\%. Given this balance, accuracy is a valid measure of performance. However, accuracy requires fixing a predicted home team win probability cutoff (typically 50\%) by which to predict a win. Given that our goal is to understand this predicted win probability, we also consider other probabilistic measures of performance.

\subsubsection{Probability measures}
We consider three metrics which directly relate the predicted probability of winning produced by each model to the actual outcome: area under the receiver operating characteristic curve (AUROC), log-loss and Brier score. We briefly explain each below.

Given a binary classifier, the receiver operating characteristic (ROC) curve plots the false positive rate (FPR) against the true positive rate (TPR). An ideal classifier would have a very high TPR for any given FPR, and thus the area under the curve would be close to one. Therefore, for AUROC, higher is better, with a maximum value of one and a minimum value of zero. Given the negative implications of predicting a team will win when they do not, AUROC is an applicable metric to measure performance of our models. 

In addition we calculate both the log-loss and the Brier score. Both statistics measure the difference between the predicted probability of a win and the actual binary value of a win/loss. Both statistics penalize the model for predictions different from the correct one. Log-loss does so by computing the cross entropy, which penalizes it in a logarthmic manner, so that small differences between predicted win probability and the actual win/loss outcome are very lightly penalized, and large differences are penalized more strongly [\cite{good1992rational}]. Brier score does so by computing the mean squared error of the difference, and thus imposes a quadratic penalty weight [\cite{brier1950verification}]. Therefore, log-loss most heavily penalizes models which get predictions very wrong, whereas Brier score quadratically increases the penalty as the difference grows. Both log-loss and Brier score have minimum values of zero, and have no finite maximum value. 

For accuracy and AUROC, higher is better. For log-loss and Brier score, lower is better.

\subsubsection{Statistical analysis}
Table \ref{tab:ModelPerfStats} summarizes the performance statistics for all models.

\begin{table} [h]
\caption{Model performance statistics}
\begin{tabular}{lcccc}
Model 	& Accuracy 	        & AUROC             & Log-Loss          & Brier score	    \\ \midrule
HomeWin & 53.15\%           & 0.5000            & 16.1820           & 0.4685            \\
LogR 	& \textbf{62.94\%} 	& \textbf{0.6768}   & \textbf{0.6418}   & \textbf{0.2254}	\\
SVM     & 62.38\%           & 0.6597            & 0.6529            & 0.2303            \\
KNN		& 62.15\%	        & 0.6686            & 0.6492            & 0.2274		    \\
XGB		& 61.08\%           & 0.6499            & 0.6639            & 0.2347            \\
ANN		& 62.13\% 	        & 0.6653            & 0.6500            & 0.2292		    \\
FTE     & 56.96\%           & 0.5950            & 0.6769            & 0.2420            \\
\end{tabular}
\label{tab:ModelPerfStats}
\end{table}

We see that all models perform better than simply predicting a home team win. We also see that, especially in terms of measures which directly relate the predicted win probabilty with the outcome, the models we described above outperform the FTE and HomeWin models.

LogR performs top across all metrics, but it is interesting to compare the metrics for other top models. We see that SVM has the second best accuracy, yet is only fourth best for Brief score and Log-loss. This is because SVM performs relatively poorly at the extremes of its prediction range, and these two metrics both penalize it for those reasons. We go into more details on this and other differences in the models in the following sections. 

Before analyzing each model in more detail, we first discuss the potential for ensembling as a way to improve predictions and also to understand the maximum possible accuracy. 

\subsubsection{Ensembling}
Based on our data and analysis, we propose that additional data and feature engineering will only result in minor improvements in accuracy. Comparing our work to that done by other authors, we see that increasing the number of columns of data collected resulted in only minor improvements in accuracy.

However, one source of improvement that is relatively untapped in this field of research is that of ensembling. By ensembling we mean training several different models such as we have done, and using their combined predictions to make a single final prediction. Ensemble models are considered state-of-the-art for many machine learning problems [\cite{sagi2018ensemble}]. The simplest form of ensembling is majority-voting, in which multiple models all make their predictions, and whichever outcome is predicted most often among them is chosen as the final prediction. One natural modification to this approach is to weight each prediction by the overall performance of the corresponding model, so that models which are known to perform well overall are given a higher weight. Yet another modification is to instead average the prediction \textit{probabilities}, so that the predicted strength of the win is taken into account. While we have described various ad-hoc approaches to ensembling, more programmatic approaches also exist [\cite{caruana2004ensemble}].

To analyze the potential for ensembling to produce stronger results, we consider both the extent to which the different models agree with one another, as well as their performance when ensembled. Indeed, the power of ensembling comes from models which are on their own highly accurate, and yet disagree with one another somewhat regularly. In Figure \ref{fig:figure_1} we see that our models agree with one another on between 81\% to 90\% of all games. As discussed above, when viewed from the viewpoint of ensembling, agreement between highly-performing models is a \textit{detriment}, as it means that when performing ensembling techniques such as majority voting, the predictions don't vary between models, so majority voting is meaningless. For this reason, we note that the relatively low level of agreement between XGB and the other models is highly desirable. Given that XGB is, on it's own, a very accuracy model, being able to ensemble it with other highly accurate models for which it has different predictions is advantageous.

\begin{figure}
    \centering
    \includegraphics[width=\linewidth]{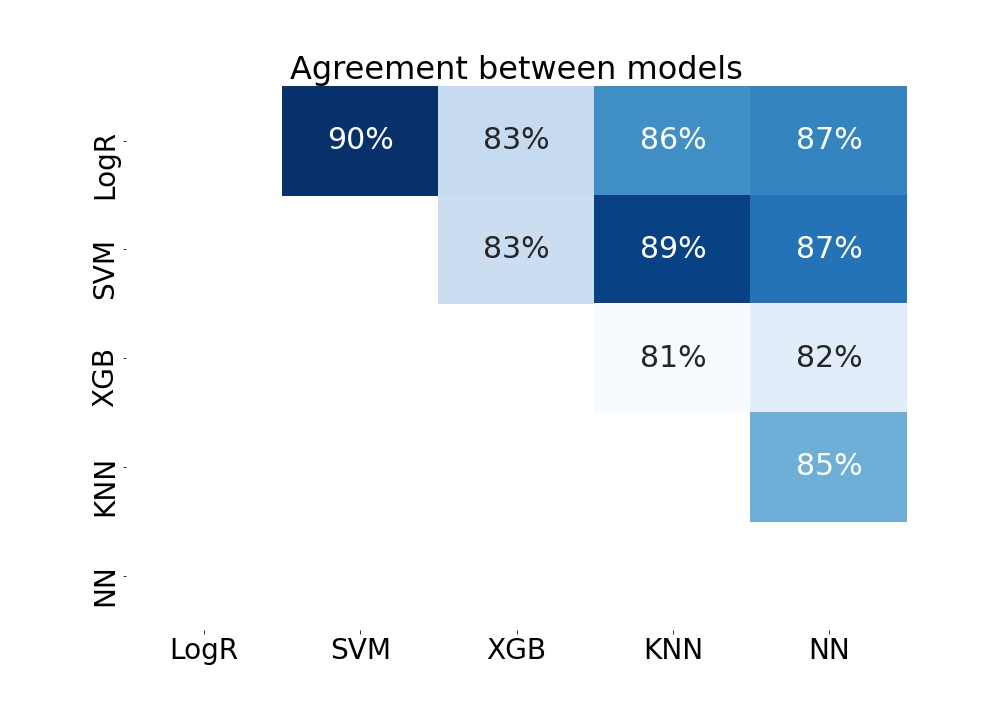}
    \caption{Percent of games on which pairs of models agree. Higher levels of agreement reduce the effectiveness of ensembling, as predictions rarely differ. Note that XGB has low agreement with all other models, and yet is highly accuracy on its own. Thus XGB is a strong candidate for model ensembling.}
    \label{fig:figure_1}
\end{figure}

In Table \ref{tab:TripletsEnsemble} we see the results of ensembling all triplets of models using majority voting. It is interesting to note that even the best ensemble models slightly under-perform the best individual model (LogR). On the other hand, even the \textit{worst} ensembled model would be the third \textit{best} model if it were only compared against the individual models. Therefore, ensembling can be seen as a form of ``smoothing'', where the effects of individual poor (or strong) predictions can potentially be dampened by the other models in the ensemble.

\begin{table} [h]
\caption{Ensembling triplets of models}
\begin{tabular}{lcccc}
Model 1 & Model 2 & Model 3 & Majority voting accuracy  & Maximum possible accuracy       \\ \midrule
LogR    & SVM     & XGB     & \textbf{62.83\%}          & 73.01\%    \\
LogR 	& SVM	  & KNN     & 62.61\%                   & 70.90\%	 \\
LogR    & SVM     & NN      & 62.63\%                   & 71.50\%    \\
LogR	& XGB	  & KNN     & 62.62\%                   & 74.10\%    \\
LogR	& XGB     & NN      & 62.60\%                   & 74.02\%    \\
LogR	& KNN 	  & NN      & 62.76\%                   & 72.47\%    \\
SVM     & XGB     & KNN     & 62.20\%                   & 73.37\%    \\
SVM     & XGB     & NN      & 62.28\%                   & 73.80\%    \\
SVM     & KNN     & NN      & 62.24\%                   & 71.58\%    \\
XGB     & KNN     & NN      & 62.28\%                   & \textbf{74.50\%}    \\
\end{tabular}
\label{tab:TripletsEnsemble}
\end{table}

\subsubsection{Maximum theoretical accuracy}
While our accuracy results are in-line with state of the art, we also wish to discuss the best possible results one could hope to obtain. Clearly, a model with perfect foresight could achieve 100\% accuracy. However, just as clearly, no such model could possibly exist. Therefore, what is the maximum possible accuracy one could hope to obtain use modern machine learning techniques and data? 

We approach this by considering the question ``what if we knew which model was correct for each game, assuming they disagree?'' If so, this would give us a theoretical maximum possible accuracy of the ensembled models. In other words, we ask ``do any models get the game in question correct?'' The results from this hypothetical experiment are shown below.

We see that the maximum accuracy from ensembling any three of these models is around 75\%. Therefore, it is perhaps feasible that with appropriate use of ensembling one could take models currently described in the literature and get prediction accuracy over 70\%. As we demonstrated above, doing so will not arise from simple majority voting. Instead, other factors must be considered. 

Finally, in the table above it is interesting to note that the greatest gains, especially in terms of maximum possible accuracy, are realized in ensembling the XGB model with essentially any other. This can also be seen by noting that XGB has a lesser agreement with the other models, while still performing at a high overall accuracy. Therefore, while XGB does not have the greatest individual accuracy, being slightly eclipsed by several other models, it shows great potential for ensembling. 

\section{Predicting win strength}
We now turn our attention to the question posed earlier: To what extent do the predicted win probabilities produced by the models correlate to score differential? For example, if team A is predicted to win with 90\% probability, does it also follow that team A will beat team B by a large margin? 

\subsection{Win probabilities}
We answer this by assessing the relationship between the predicted win probability produced by each model and score differential. The primary questions we consider are:
\begin{enumerate}
    \item To what extent do models show a positive relationship between predicted win probability and strength of a win?
    \item To what extent are toss-up games correctly identified by the models?
    \item If a dominant win is predicted, does that correlate to a large final score differential?
\end{enumerate}

While questions 2 and 3 are sub-questions of question 1, they are important questions in their own right. For betting purposes, question 2 is especially important since run-line betting requires winning by a sufficient margin [\cite{betmgm_runline_2022}], as well as understanding how well models understand score differential as a measure of strength of a win. Question 3 is important as it may help teams influence their choice of strategies for games in which a high probability of wining (or losing) is predicted. 

All models show a positive relationship between predicted probabilities and score differential. While the individual $R^2$ values remain around 0.1, the random nature of sports dictate that predictions can never, on a game-by-game basis approach perfection. However, the general trends show that on average, all models seem to gain an intuitive understanding of ``winning'' as more than just ``having a score higher than the other team.'' We summarize the relationship between predicted home team win probability and home team score differential (defined as home team final score minus away team final score) in Figure \ref{fig:figure_2}. These suggest a positive answer to question 1, as most models do indeed show a positive linear relationship between predicted win probability and score differential. The best performing models, such as LogR, KNN and ANN all demonstrate clear relationships.

\begin{figure}[!h]
\begin{subfigure}{.5\textwidth}
  \centering
  \includegraphics[]{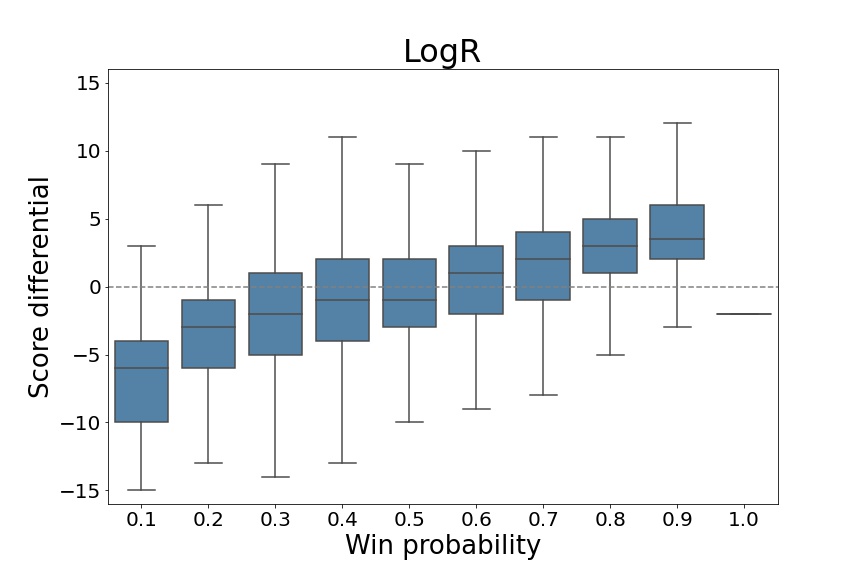}
\end{subfigure}%
\begin{subfigure}{.5\textwidth}
  \centering
      \includegraphics[]{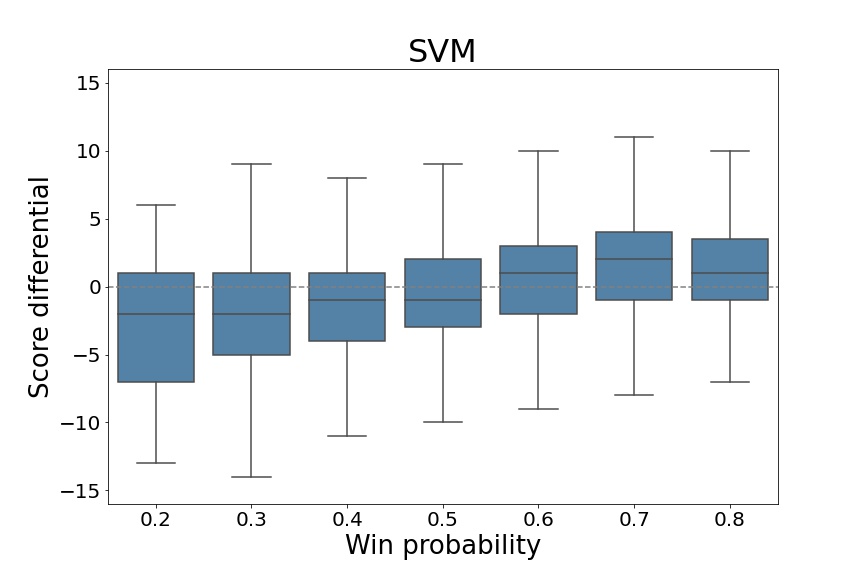}
\end{subfigure}
\begin{subfigure}{.5\textwidth}
  \centering
      \includegraphics[]{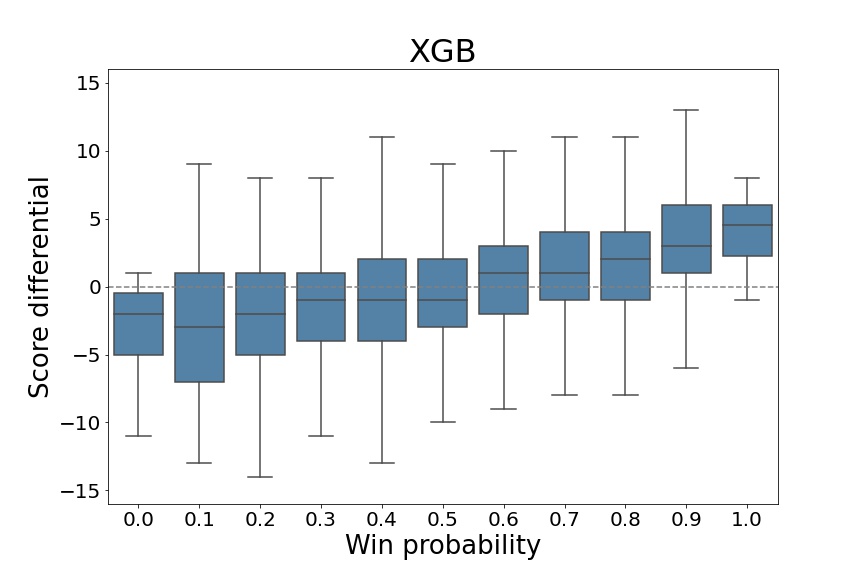}
\end{subfigure}%
\begin{subfigure}{.5\textwidth}
  \centering
      \includegraphics[]{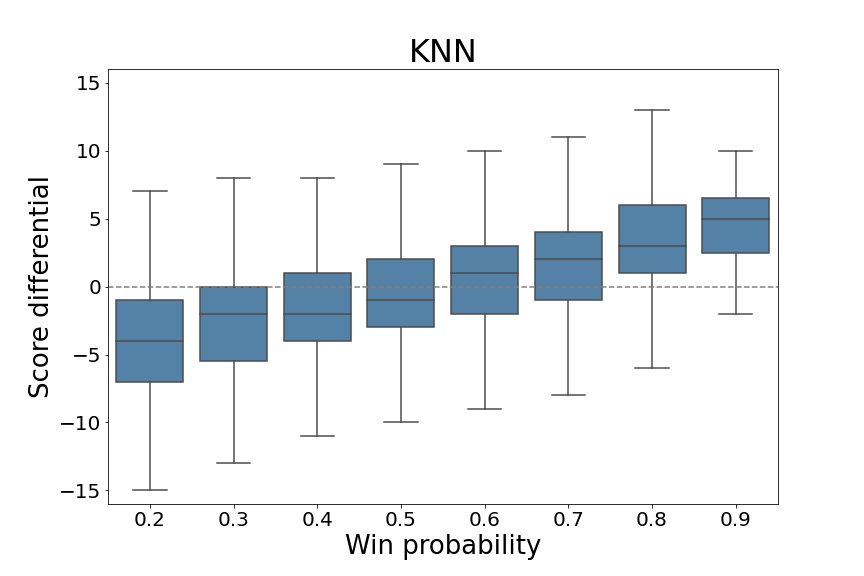}
\end{subfigure}
\begin{subfigure}{.5\textwidth}
  \centering
      \includegraphics[]{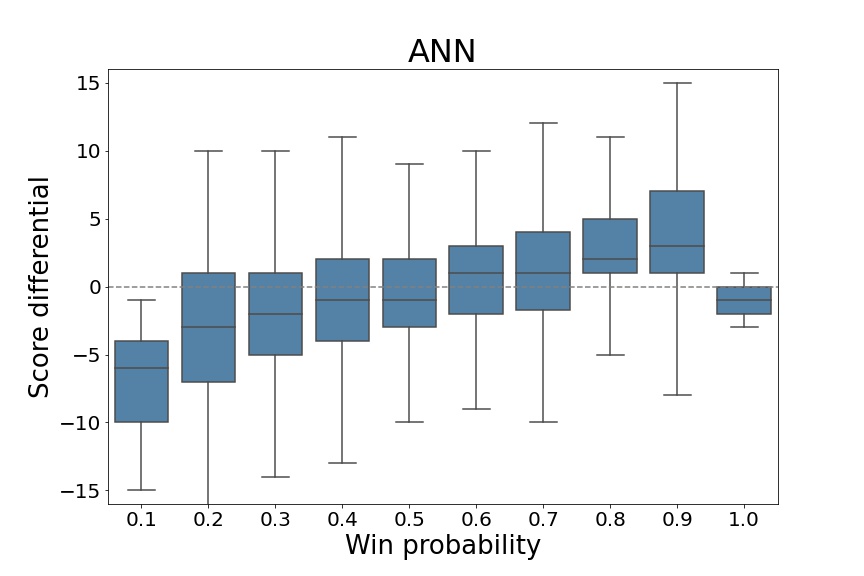}
\end{subfigure}%
\begin{subfigure}{.5\textwidth}
  \centering
      \includegraphics[]{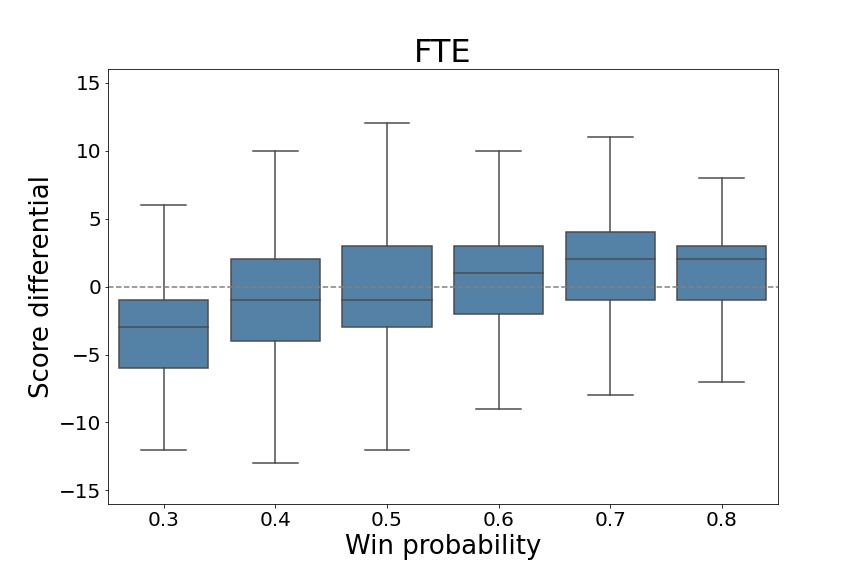}
\end{subfigure}
\caption{Predicted home team win probability versus score differential (home team final score minus away team final score). All predicted win probabilities are unscaled and rounded to the nearest 10\%. A dashed horizontal line at zero score differential (tied game) is shown for reference. LogR, KNN and ANN show the strongest positive linear trends. XGB and KNN perform the best for games with the most extreme predicted win probabilities. Note that SVM, KNN and FTE all fail to predict any games to have an especially high or low win probability.}
\label{fig:figure_2}
\end{figure}

\begin{table}[h]
\caption{Probability relationship statistics}
\begin{tabular}{lcc}
Model 	& $R^2$ with score differential   \\ \midrule
FTE     & 0.036                                       \\
LogR 	& \textbf{0.140} 	             	    \\
SVM     & 0.105 \\
KNN		& 0.063	                            		\\
XGB		& 0.098                                     \\
ANN		& 0.120  		\\
\end{tabular}
\label{tab:ModelR2}
\end{table}

To answer the second question we considered only games in which each model predicted a probability of a home team win near 50\%. We refer to these as ``toss-up games'', and consider two possible ranges of probabilities, namely $50\% \pm 5\%$ and $50\% \pm 1\%$. In table \ref{tab:TossUp} we see that all models which have predicted win probabilities near 50\% do indeed correlate with close games. In all but one model (XGB) for just one prediction range (49\%-51\%), the average score differential remained less than 0.5. We do note however that the standard deviation for the corresponding score differentials for each model are all above four, which is relatively large. This means that, on average, we may conclude that toss-up games are well-captured, but for any individual game there is still considerable variation. This helps explain the low $R^2$ values from Table \ref{tab:ModelR2}. It is also interesting to note that even when the prediction range narrows and by definition less games are included in the calculation, the standard deviation for all models remains relatively unchanged. This indicates that, even though fewer games were considered, the reliability of estimates increased somewhat, which offsets the increase in standard deviation coming from reducing the sample size.

Note that all mean score differentials are negative, even though the range of predicted win probabilities are symmetric around 50\%. This means that when a model predicts a toss-up game, there is actually slightly more chance of the home team losing than winning. Ideally all models would predict mean score differentials of zero. These results suggest that, when dealing with toss-up games in particular, ANN may be the best choice, as its mean score differentials are closest to zero, and its standard deviations are quite small.

\begin{table}[h]
\caption{Toss-up statistics}
\begin{tabular}{lccc}
Model 	& Prediction range  & Number of games & Mean score differential ($\pm$ sd)       \\ \midrule
FTE & 45\%-55\%           & 4471            & -0.227 $\pm$ 4.422                   \\
 & 49\%-51\%           & 1026             & -0.280 $\pm$ 4.500      \\ \\  
LogR & 45\%-55\%           & 2289            & -0.371 $\pm$ 4.120                  \\
 & 49\%-51\%           & 461             & -0.243 $\pm$ 4.011                     \\\\
SVM & 45\%-55\%           & 2303            & -0.370 $\pm$ 4.126                  \\
 & 49\%-51\%           & 462             & -0.240 $\pm$ 4.007                     \\\\
KNN & 45\%-55\%           & 3396             & -0.158 $\pm$ 4.502    \\
 & 49\%-51\%           & 677             & -0.233  $\pm$ 4.505                     \\\\
XGB & 45\%-55\%           & 1996            & -0.333 $\pm$ 4.090   \\
 & 49\%-51\%           & 415             & -0.572 $\pm$ 4.505                  \\\\
ANN & 45\%-55\%           & 1926             & -0.391 $\pm$ 4.189                     \\
 & 49\%-51\%           & 426             & -0.153 $\pm$ 4.089                    \\
\end{tabular}
\label{tab:TossUp}
\end{table}

Finally we address the third question. We see in table \ref{tab:HomeFav} that predicted dominant wins resulted in average score differentials of at least two runs, and closer to four runs for the best models. Similarly for predicted dominant away team wins in Table \ref{tab:HomeUnder} we see an average score differential of at least two runs and up to six or more runs for the best models. It is interesting to note that the standard deviations for these dominant wins/losses are not much larger than the corresponding standard deviations for toss-up games. This is meaningful because one may expect that for a model predicting a strong win, a broad range of possible outcomes are all equally likely (winning by anything from one run to ten or more). However this suggests that this is not the case. It is also noteworthy that the SVM and KNN models have a more concentrated range of predicted win probabilities, leading to no games with predicted wins in the highest and lowest prediction range. This is a disadvantage compared to the other models, as having a broad range of predicted win probabilities allows for one to more easily differentiate ``sure wins'' from closer games.

\begin{table}[h]
\caption{Home favorite predictions}
\begin{tabular}{lccc}
Model 	& Prediction range  & Number of games & Mean score differential ($\pm$ sd)      \\ \midrule
FTE & 75\%-100\% & 9 & 2.667 $\pm$ 6.423  \\
 & 85\%-100\% & 0 & -  \\ \\
LogR & 75\%-100\%           & 1095            & 3.069 $\pm$ 4.152                      \\
 & 85\%-100\%           & 181             & 4.072 $\pm$ 3.978   \\\\
SVM & 75\%-100\%           & 43            & 1.419 $\pm$ 4.452                  \\
 & 85\%-100\%           & 0             & -                    \\\\
KNN & 75\%-100\%           & 94             & 2.5 $\pm$ 3.423      \\
 & 85\%-100\%           & 0             & -                    \\\\
XGB & 75\%-100\%           & 1185            & 2.388 $\pm$ 4.373                    \\
 & 85\%-100\%           & 334             & 3.434 $\pm$ 4.151                     \\\\
ANN & 75\%-100\%           & 1504             & 2.557 $\pm$ 4.167                    \\
 & 85\%-100\%           & 341             & 3.754 $\pm$ 4.455                    \\
\end{tabular}
\label{tab:HomeFav}
\end{table}

\begin{table}[h]
\caption{Home underdog predictions}
\begin{tabular}{lccc}
Model 	& Prediction range  & Number of games & Mean score differential ($\pm$ sd) 	    \\ \midrule
FTE & 0\%-25\% & 0 & -  \\
 & 0\%-15\% & 0 & -  \\ \\
LogR & 0\%-25\%           & 327            & -3.651 $\pm$ 4.533                     \\
 & 0\%-15\%           & 25             & -6.920 $\pm$ 5.507                     \\\\
SVM & 0\%-25\%           & 75            & -2.453 $\pm$ 4.731                  \\
 & 0\%-15\%           & 0             & -                     \\\\
KNN & 0\%-25\%           & 0            & -                       \\
 & 0\%-15\%           & 0             & -                      \\\\
XGB & 0\%-25\%           & 671            & -2.575 $\pm$ 4.467                      \\
 & 0\%-15\%           & 137             & -3.161 $\pm$ 4.626                      \\\\
ANN & 0\%-25\%           & 201            & -4.224 $\pm$ 5.109                      \\
 & 0\%-15\%           & 6             & -6.500 $\pm$ 3.619                    \\\\
\end{tabular}
\label{tab:HomeUnder}
\end{table}

\subsection{FTE model predictions}
We next show the general trend of the predictions of the FiveThirtyEight model discussed earlier. As seen in Figure \ref{fig:figure_3}, their predictions tend to stay nearer to 50\%, with a 25th percentile of 48.3\% predicted win probability and a 75th percentile of 58.7\%. Thus the majority of their predictions stay quite close to 50\%. In Figure \ref{fig:figure_3} we contrast these with predicted win probabilities from our LogR model.

\begin{figure}[!h]
\includegraphics[width=\linewidth]{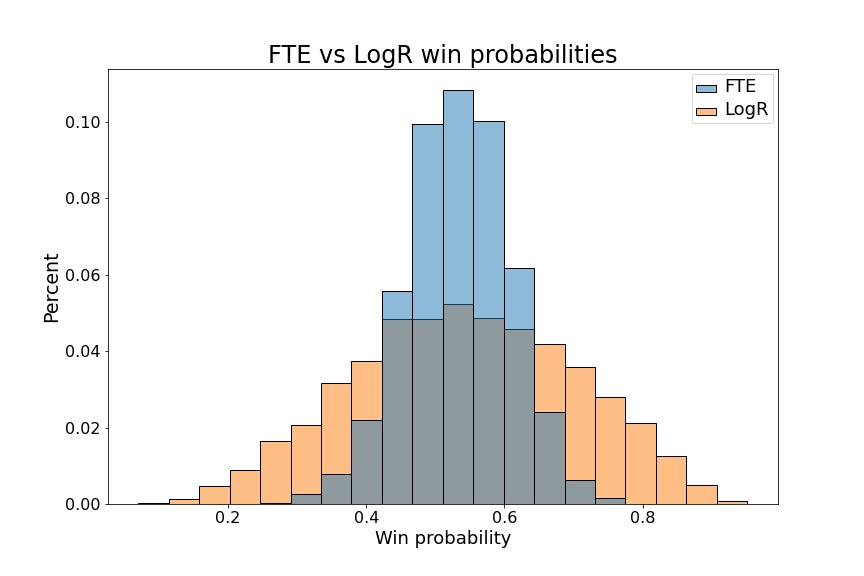}
\caption{Distribution of predicted home team win probabilities for FTE versus LogR. Note the narrow prediction region for FTE and the skinny tails. This means that many games will marked as being toss-ups.}
\label{fig:figure_3}
\end{figure}

Of course, any model can be scaled to have predictions in an arbitrary range. Yet even after scaling our model shows a noticeably broader distribution of predicted win probabilities due to the fat tails of the LogR distribution compared to FTE. In addition, in order to scale predictions, \textit{all predictions made by that model} must be known ahead of time. This is often prohibitive, and means that, even though one prediction may have been scaled, it may need to be re-scaled again in the future.

\section{Application to run-line betting}
Betting in baseball typically takes one of three forms: moneyline, over/under and run-line. Moneyline bets simply attempt to choose the winner of the game, with the payouts set by the odds maker. Over/under bets on the total number of points in the game, regardless of who wins. Run-line betting also picks a winner, but the winner must win by a sufficiently large amount (usually 1.5 runs). Given the focus of our research, we focus on run-line betting, as it rewards bettors for choosing not only the winning team, but the team knowing that the team will win by a sufficient amount to cover the run line. However, our analysis applies equally well to moneyline betting, and could easily be modified to analyze over/under betting.

\subsection{Historical betting data}
We source historical betting data from Sports Books Reviews Online [\cite{sportsbooksreviewsonline}] which sources data from various offshore and Nevada land based sports books. This data includes run-lines and run-line money for each game on a season-by-season basis.

\subsection{Naive model}
We first consider the naive model which bets on the home team to win when the predicted win probability is 0.5 or greater, and on the away team otherwise. We use the LogR model for thes predictions as it has the best overall performance. However, following this naive betting strategy has poor performance, giving a loss of 51.39\% of money wagered. 

The primary issue with this naive model is that every game is bet on. However, for many games where the model has a predicted probability of winning near 0.5, it makes more sense to simply \textit{not} bet on the game. In those cases one would be better off abstaining from betting on the game and use that money for a more ``sure bet.'' Indeed, as we showed earlier, there is a clear linear relationship between the predicted probability of winning and the score differential. Therefore games with a predicted probability of winning near 0.5 tend to be close games that perhaps could have swung either direction, or for whom the score differential would not have exceeded the run-line as demonstrated in Table \ref{tab:TossUp}. To remedy this, we optimize our betting model by allowing some games to not be bet on.

\subsection{Optimized model}
In order to optimize our betting model we instead choose two different probability cutoffs. One cutoff is sufficiently high so that any game with a predicted probability higher than that will be predicted to be a win, and similarly a low cutoff indicating a loss. Any game with a predicted win probability between the high and low cutoffs is not bet on. Note that this reduces to the naive model when both the high and low cutoff equal 0.5. However, when optimizing the model we explore whether allowing these two cutoffs to diverge will result in better returns. 

To analyze this, we chose twenty equally spaced low cutoffs between 0 and 0.5, and twenty high cutoffs between 0.5 and 1. We then evaluate the performance with all four hundred possible pairs of cutoffs. We use the best performing model (LogR) for predictions. We evaluate the results on the 2016-2019 season. We chose these seasons because, since the models were trained on the 2001-2015 seasons, it would be a methodological error to evaluate their performance on the same seasons.

Running this analysis shows that it is indeed possible to choose cutoffs which give positive returns. In Figure \ref{fig:figure_4} we show the returns associated with each pair of cutoffs. We calculate returns as a percentage of money invested.  Note that when the high and low cutoffs differ, there is an area between the low and high cutoff for which games are not wagered on. This is key to generating positive returns.

Another concern is the number of games wagered on. While large returns are desirable, the cutoffs chosen can be highly risky if it results in only betting large amounts of money on only a small handful of games. For this reason we also report what percentage of games were wagered on. Recall that each of the 30 teams play 162 regular season games, giving a total of 2430 games per season. In Figure \ref{fig:figure_5} we report the percentage of games bet on per season with the color corresponding to the returns shown in Figure \ref{fig:figure_4}. We see that in the areas with positive returns approximately 0.5\% to 5\% of games are bet on. This corresponds to about 5 to 120 games per season, with the highest returns being realized at around 1\% to 2\% of games wagered on, or roughly 25 to 50 games per season.

\begin{figure}[!h]
\centering
\includegraphics[width=\linewidth]{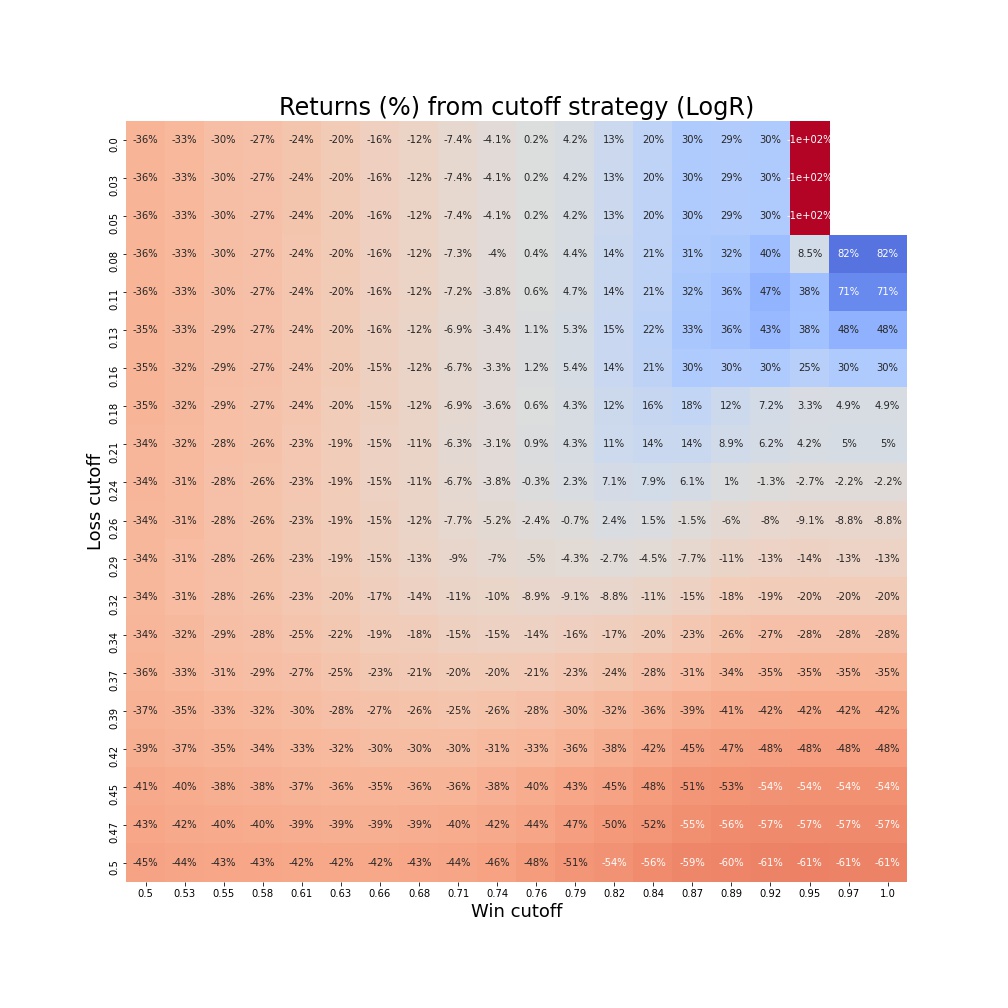}
\caption{Returns as a percentage of money invested using the high and low cutoffs shown on each axis. Predictions come from the LogR model. Setting the low and high cutoff equal to 0.5 is precisely the naive strategy described above. For appropriate choices of win and loss cutoffs we demonstrate positive returns, even into the double digits.}
\label{fig:figure_4}
\end{figure}

\begin{figure}[!h]
\centering
\includegraphics[width=\linewidth]{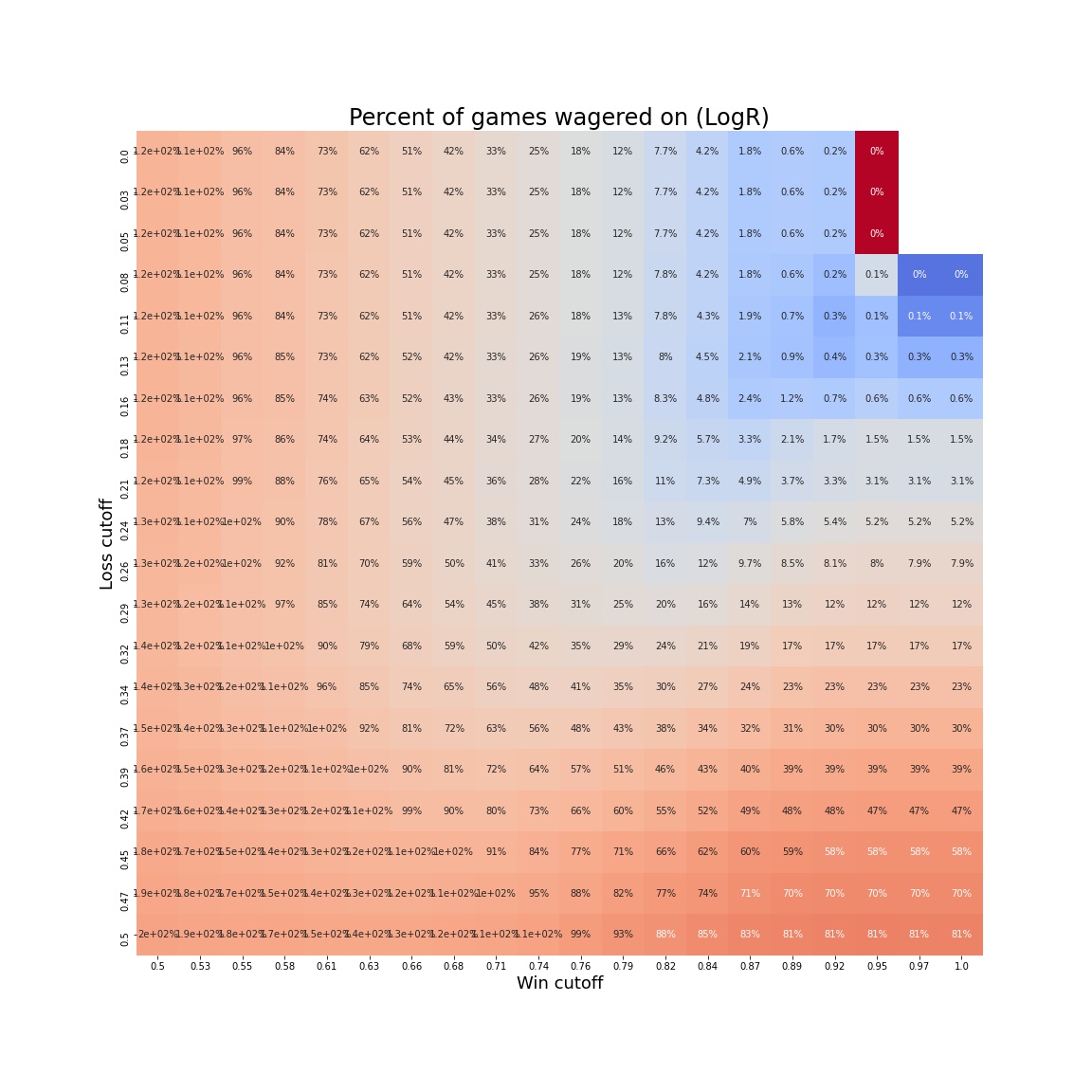}
\caption{Percentage of games wagered on using each pair of cutoffs using the LogR model. Colors correspond to the returns shown in the previous figure. Note that in situations where positive returns are realized, approximately 0.5\% to 5\% of games are wagered on. In a normal season of 2430 games this corresponds to between 5 and 120 games per season.}
\label{fig:figure_5}
\end{figure}

\section{Conclusions}
While other researchers have attempted to predict wins, none have considered the strength of a win. Rather than modeling the strength directly, we model wins using a binary classifier, but show that the models indeed gain a deep understanding of win probability. We show the applicability of these probabilities to score differential. Our analysis shows that nearly any model, when trained with appropriate and sufficient data, will produce predicted win probabilities that correlate with score differential. Indeed, we believe that the differences between them were generally small enough that no one model should be considered ``best.'' However, if interpretability is an important factor, then logistic regression should be a first choice, with K-nearest neighbors also being a strong candidate. If, on the other hand, the ability to fine-tune the model matters more, perhaps at the expense of interpretability, then a gradient boosting model is an excellent choice, along with a feed-forward neural network. We also see that every model significantly outperforms the baseline model of predicting a home team win.

\subsection{Future work}
Our work showed that positive run-line betting returns are possible using an approach of independently tuning the low and high probability cutoffs. However, a similar approach could consider weighting the amount wagered by the probability of winning. In that case, one is changing from using a hard cutoff to one in which the cutoff and amount to be wagered are related. It is reasonable to assume that returns will improve under that approach. 

We also suggest analyzing drawdown, which is a temporal analysis of investment returns. Drawdown is defined as the maximum return at any point in the season minus the minimum return [\cite{investo_drawdown}]. A large drawdown indicates wild swings in returns throughout the season, resulting in higher risk for the bettor. We hypothesize that the weighted approach described in this section would potentially result in lower drawdown than the approach we described in the run-line betting section above.

While the accuracy we achieved is in-line with state of the art, with single models around 63\% accuracy, and ensembled models around 70\%, our experiments around ensembling show that theoretical maximum accuracy is around 75\%. Therefore, we suggest more work could be done on alternative ensembling techniques. 

\begin{acknowledgement}
  We thank the St. Edward's University Department of Mathematics for their support.
\end{acknowledgement}

\bibliographystyle{agsm}
\bibliography{bibliography}

@book{menard:2002,
    author =        {Menard, Scott},
    year =          {2002},
    title =         {Applied logistic regression analysis},
    publisher =     {Sage},
    volume =        {106}
}

@article{dreiseitl:2002,
    author = "Dreiseitl, Stephan,  Lucila Ohno-Machado",
    title = "Logistic regression and artificial neural network classification models: a methodology review.",
    journal = "Journal of biomedical informatics",
    volume = "35",
    number = "5--6",
    pages = "352--359",
    year = "2002",
    DOI = ""
}

@article{laaksonen:1996,
    author = "Laaksonen, Jorma,  Erkki Oja",
    title = "Classification with learning k-nearest neighbors",
    journal = "Proceedings of International Conference on Neural Networks",
    volume = "3",
    number = "",
    pages = "1480--1483",
    year = "1996",
    DOI = ""
}

@article{chen:2016,
    author = "Chen, Tianqi,  Carlos Guestrin",
    title = "Xgboost: A scalable tree boosting system",
    journal = "Proceedings of the 22nd ACM SIGKDD international conference on knowledge discovery and data mining",
    volume = "",
    number = "",
    pages = "785--794",
    year = "2016",
    DOI = ""
}

@article{baboota:2019,
    author = "Baboota, Rahul,  Harleen Kaur",
    title = "Predictive analysis and modelling football results using machine learning approach for English Premier League",
    journal = "International Journal of Forecasting",
    volume = "35",
    number = "2",
    pages = "741--755",
    year = "2019",
    DOI = ""
}

@article{alfredo:2019,
    author = "Alfredo, Yoel F.,  Sani M. Isa",
    title = "Football match prediction with tree based model classification",
    journal = "IJ Intelligent Systems and Applications",
    volume = "11",
    number = "7",
    pages = "20--28",
    year = "2019",
    DOI = ""
}

@article{ranganathan:2017,
    author = "Ranganathan, Priya and C. S. Pramesh and Rakesh Aggarwal",
    title = "Common pitfalls in statistical analysis: logistic regression",
    journal = "Perspectives in clinical research",
    volume = "8",
    number = "3",
    pages = "148",
    year = "2017",
    DOI = ""
}

@article{li:2015,
    author = "Li, Jiwei and Xinlei Chen and Eduard Hovy and Dan Jurafsky",
    title = "Visualizing and understanding neural models in nlp",
    journal = "arXiv preprint",
    volume = "",
    number = "",
    pages = "",
    year = "2015",
    DOI = ""
}

@article{borisov:2021,
    author = "Borisov, Vadim and Tobias Leemann and Kathrin Seßler and Johannes Haug and Martin Pawelczyk and Gjergji Kasneci",
    title = "Deep neural networks and tabular data: A survey",
    journal = "arXiv preprint",
    year = "2021",
    DOI = "arXiv:2110.01889"
}

@article{jurgen:2001,
    author = "Perl, Jürgen",
    title = "Artificial neural networks in sports: New concepts and approaches",
    journal = "International Journal of Performance Analysis in Sport 1",
    volume = "",
    number = "1",
    pages = "106--121",
    year = "2001",
    DOI = ""
}

@article{jurgen:2003,
    author = "Jürgen, Perl,  Baca Arnold",
    title = "Application of neural networks to analyze performance in sports",
    journal = "Proceedings of the 8th Annual Congress of the European College of Sport Science",
    volume = "342",
    number = "",
    pages = "",
    year = "2003",
    DOI = ""
}

@article{deselits:1992,
    author = "DeSilets, L. and B. Golden and R. Kumar and Q. Wang",
    title = "A neural network model for cell suppression of tabular data",
    journal = "IJCNN International Joint Conference on Neural Networks",
    volume = "3",
    number = "",
    pages = "203--214",
    year = "1992",
    DOI = ""
}

@book{khan:2018,
    author = "Khan, Salman and Hossein Rahmani and Syed Afaq Ali Shah and Mohammed Bennamoun",
    title = "A guide to convolutional neural networks for computer vision",
    journal = "Synthesis Lectures on Computer Vision 8",
    pages = "1--207",
    year = "2018",
    publisher = ""
}

@article{hornik:1989,
    author = "Hornik, Kurt and Maxwell Stinchcombe and Halbert White",
    title = "Multilayer feedforward networks are universal approximators",
    journal = "Neural networks 2",
    volume = "",
    number = "5",
    pages = "359--366",
    year = "1989",
    DOI = ""
}

@book{gurney:2018,
    author = "Gurney, Kevin",
    title = "An introduction to neural networks",
    publisher = "CRC press",
    year = "2018"
}

@book{best:2015,
    author = "Best, Henning,  Christof Wolf",
    title = "The SAGE handbook of regression analysis and causal inference",
    chapter = "Logistic regression",
    publisher = "",
    year = "2015",
    pages = "153--172"
}

@misc{fte:2022,
  title = "How Our MLB Predictions Work",
  author = {Boice, Jay and Silver, Nate},
  year={2022},
  howpublished = {\url{https://fivethirtyeight.com/methodology/how-our-mlb-predictions-work/}},
  note = {Accessed: 2022-01-19}
}

@article{james1981baseball,
  title={Baseball Abstract, self-published},
  author={James, Bill},
  journal={Lawrence, KS},
  year={1981}
}

@article{barrow2013ranking,
  title={Ranking rankings: an empirical comparison of the predictive power of sports ranking methods},
  author={Barrow, Daniel and Drayer, Ian and Elliott, Peter and Gaut, Garren and Osting, Braxton},
  journal={Journal of Quantitative Analysis in Sports},
  volume={9},
  number={2},
  pages={187--202},
  year={2013},
  publisher={De Gruyter}
}

@article{miller2007derivation,
  title={A derivation of the pythagorean won-loss formula in baseball},
  author={Miller, Steven J},
  journal={Chance},
  volume={20},
  number={1},
  pages={40--48},
  year={2007},
  publisher={Taylor \& Francis}
}

@article{jia2013predicting,
  title={Predicting the Major League Baseball Season},
  author={Jia, Randy and Wong, Chris and Zeng, David},
  journal={CS229 Machine Learning Final Project},
  pages={1--5},
  year={2013}
}

@article{soto2016predicting,
  title={Predicting Win-Loss outcomes in MLB regular season games--A comparative study using data mining methods},
  author={Soto Valero, C{\'e}sar},
  journal={International Journal of Computer Science in Sport},
  volume={15},
  number={2},
  year={2016}
}

@article{huang2021use,
  title={Use of Machine Learning and Deep Learning to Predict the Outcomes of Major League Baseball Matches},
  author={Huang, Mei-Ling and Li, Yun-Zhi},
  journal={Applied Sciences},
  volume={11},
  number={10},
  pages={4499},
  year={2021},
  publisher={Multidisciplinary Digital Publishing Institute}
}

@article{elfrink2018predicting,
  title={Predicting the outcomes of MLB games with a machine learning approach},
  author={Elfrink, Tim},
  journal={Vrije Universiteit Amsterdam},
  year={2018}
}

@article{beneventano2012predicting,
  title={Predicting run production and run prevention in baseball: the impact of Sabermetrics},
  author={Beneventano, Philip and Berger, Paul D and Weinberg, Bruce D},
  journal={International Journal of Business, Humanities and Technology},
  volume={2},
  number={4},
  pages={67--75},
  year={2012}
}

@article{bergmeir2012use,
  title={On the use of cross-validation for time series predictor evaluation},
  author={Bergmeir, Christoph and Ben{\'\i}tez, Jos{\'e} M},
  journal={Information Sciences},
  volume={191},
  pages={192--213},
  year={2012},
  publisher={Elsevier}
}

@article{chu2019empirical,
  title={Empirical study on relationship between sports analytics and success in regular season and postseason in Major League Baseball},
  author={Chu, David P and Wang, Cheng W},
  journal={Journal of Sports Analytics},
  volume={5},
  number={3},
  pages={205--222},
  year={2019},
  publisher={IOS Press}
}

@article{conforti2021analysis,
  title={An Analysis of Playoff Performance Declines in Major League Baseball},
  author={Conforti, Christian M and Crotin, Ryan L and Oseguera, Jordan},
  journal={The Journal of Strength \& Conditioning Research},
  volume={35},
  pages={S36--S41},
  year={2021},
  publisher={LWW}
}

@article{jones2015home,
  title={The home advantage in major league baseball},
  author={Jones, Marshall B},
  journal={Perceptual and motor skills},
  volume={121},
  number={3},
  pages={791--804},
  year={2015},
  publisher={SAGE Publications Sage CA: Los Angeles, CA}
}

@article{stefani2008measurement,
  title={Measurement and interpretation of home advantage},
  author={Stefani, Ray},
  journal={Statistical thinking in sports},
  pages={203--216},
  year={2008},
  publisher={Chapman and Hall/CRC New York}
}

@book{hastie2009elements,
  title={The elements of statistical learning: data mining, inference,  prediction},
  author={Hastie, Trevor and Tibshirani, Robert and Friedman, Jerome H and Friedman, Jerome H},
  volume={2},
  year={2009},
  pages={417--421},
  publisher={Springer}
}

@article{sagi2018ensemble,
  title={Ensemble learning: A survey},
  author={Sagi, Omer and Rokach, Lior},
  journal={Wiley Interdisciplinary Reviews: Data Mining and Knowledge Discovery},
  volume={8},
  number={4},
  pages={e1249},
  year={2018},
  publisher={Wiley Online Library}
}

@inproceedings{caruana2004ensemble,
  title={Ensemble selection from libraries of models},
  author={Caruana, Rich and Niculescu-Mizil, Alexandru and Crew, Geoff and Ksikes, Alex},
  booktitle={Proceedings of the twenty-first international conference on Machine learning},
  pages={18},
  year={2004}
}

@misc{betmgm_runline_2022, 
    title={What is a run line in MLB betting?}, url={https://sports.betmgm.com/en/blog/baseball-what-is-a-run-line-mlb-betting-odds/}, 
    journal={BetMGM}, 
    author={BetMGM}, 
    year={2022}, 
    month={Apr}
}

@incollection{good1992rational,
  title={Rational decisions},
  author={Good, Irving John},
  booktitle={Breakthroughs in statistics},
  pages={365--377},
  year={1992},
  publisher={Springer}
}

@article{brier1950verification,
  title={Verification of forecasts expressed in terms of probability},
  author={Brier, Glenn W and others},
  journal={Monthly weather review},
  volume={78},
  number={1},
  pages={1--3},
  year={1950}
}

@article{healey2017new,
  title={The new Moneyball: How ballpark sensors are changing baseball},
  author={Healey, Glenn},
  journal={Proceedings of the IEEE},
  volume={105},
  number={11},
  pages={1999--2002},
  year={2017},
  publisher={IEEE}
}

@article{lage2016statcast,
  title={Statcast dashboard: Exploration of spatiotemporal baseball data},
  author={Lage, Marcos and Ono, Jorge Piazentin and Cervone, Daniel and Chiang, Justin and Dietrich, Carlos and Silva, Claudio T},
  journal={IEEE computer graphics and applications},
  volume={36},
  number={5},
  pages={28--37},
  year={2016},
  publisher={IEEE}
}

@article{davenport2014businesses,
  title={What businesses can learn from sports analytics},
  author={Davenport, Thomas H},
  journal={MIT Sloan Management Review},
  volume={55},
  number={4},
  pages={10},
  year={2014},
  publisher={Massachusetts Institute of Technology, Cambridge, MA}
}

@article{elitzur2020data,
  title={Data analytics effects in major league baseball},
  author={Elitzur, Ramy},
  journal={Omega},
  volume={90},
  pages={102001},
  year={2020},
  publisher={Elsevier}
}

@article{zhang2021survey,
  title={A survey on neural network interpretability},
  author={Zhang, Yu and Ti{\v{n}}o, Peter and Leonardis, Ale{\v{s}} and Tang, Ke},
  journal={IEEE Transactions on Emerging Topics in Computational Intelligence},
  year={2021},
  publisher={IEEE}
}

@misc{fte,
    title={FiveThirytEight},
    url={https://fivethirtyeight.com/},
    author={Silver, Nate},
    journal={BetMGM}, 
    year={2022}, 
    month={Apr}
}

@misc{sportsbooksreviewsonline,
    title={Sports Books Reviews Online},
    url={https://www.sportsbookreviewsonline.com/scoresoddsarchives/scoresoddsarchives.htm},
    author={SBRO},
    journal={Sportbooks Reviews Online}, 
    year={2022}, 
    month={Apr}
}

@misc{investo_drawdown,
    title={Drawdown Definition},
    url={https://www.investopedia.com/terms/d/drawdown.asp},
    author={Mitchell, Cory},
    journal={Investopedia},
    year={2022},
    month={March}
}

@misc{baseball_reference,
    title={Major League Players},
    url={https://www.baseball-reference.com/players/},
    author={SportsReferenceLLC},
    journal={Baseball Reference},
    year={2022},
    month={Apr}
}

@misc{fangraphs,
    title={Baseball Data Tools},
    url={https://www.fangraphs.com},
    author={FanGraphs},
    journal={FanGraphs},
    year={2022},
    month={Apr}
}

@misc{lahman,
    title={Lahman's Baseball Database},
    url={https://www.seanlahman.com/baseball-archive/statistics/},
    author={Lahman, Sean},
    journal={www.SeanLahman.com},
    year={2022},
    month={Apr}
}

@misc{retrosheet,
    title={Retrosheet},
    url={https://www.retrosheet.org},
    author={Smith, Dave},
    journal={Retrosheet},
    year={2022},
    month={Apr}
}

@misc{fte_github_mlb_elo,
    title={MLB ELO},
    url={https://github.com/fivethirtyeight/data/tree/master/mlb-elo},
    author={Silver, Nate},
    journal={Github},
    year={2022},
    month={Apr}
}

\newpage

\begin{figure}
    \centering
    \includegraphics[width=\linewidth]{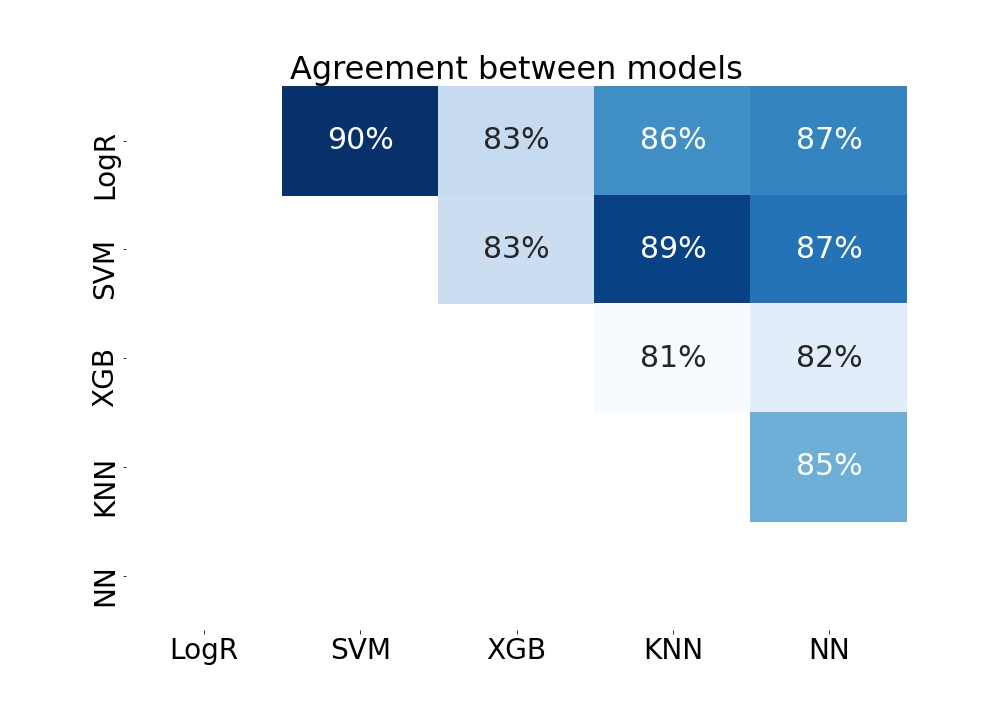}
    \caption{Percent of games on which pairs of models agree. Higher levels of agreement reduce the effectiveness of ensembling, as predictions rarely differ. Note that XGB has low agreement with all other models, and yet is highly accuracy on its own. Thus XGB is a strong candidate for model ensembling.}
    \label{fig:figure_1}
\end{figure}

\newpage

\begin{figure}[!h]
\begin{subfigure}{.5\textwidth}
  \centering
  \includegraphics[width=\linewidth]{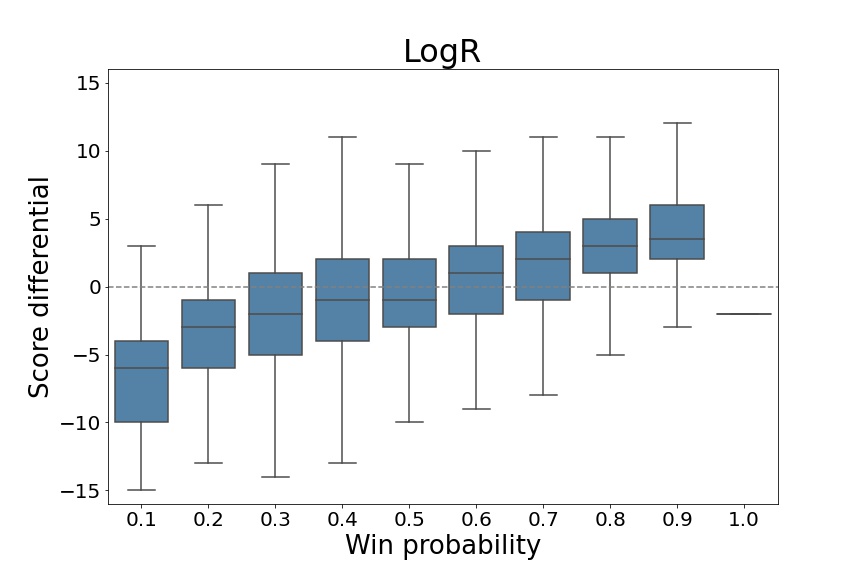}
\end{subfigure}%
\begin{subfigure}{.5\textwidth}
  \centering
      \includegraphics[width=\linewidth]{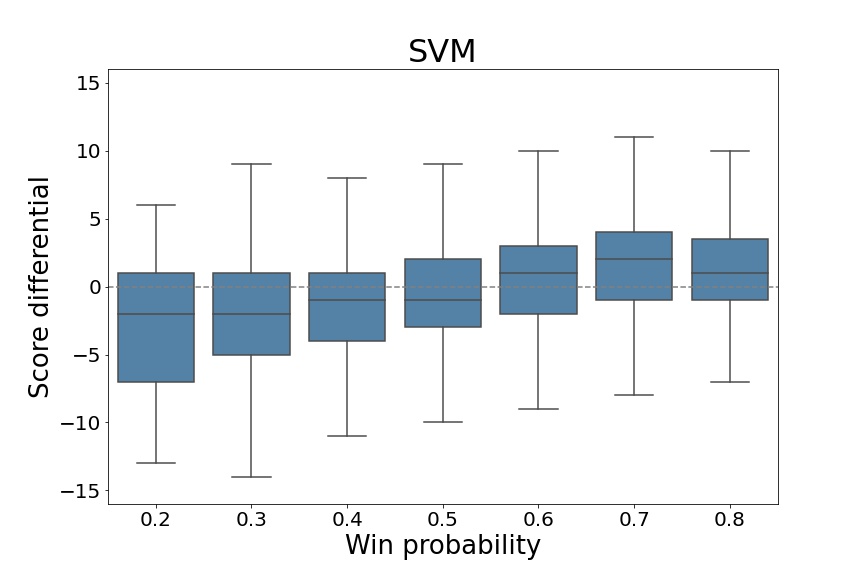}
\end{subfigure}
\begin{subfigure}{.5\textwidth}
  \centering
      \includegraphics[width=\linewidth]{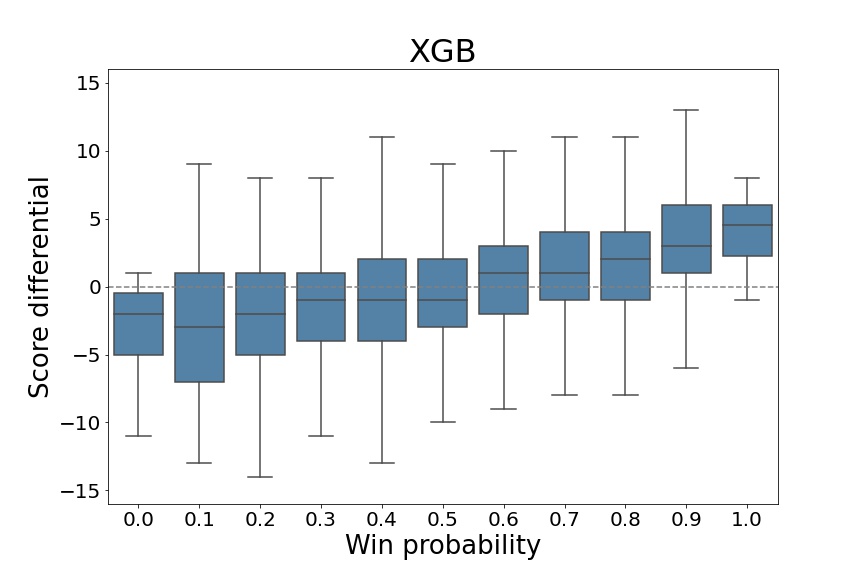}
\end{subfigure}%
\begin{subfigure}{.5\textwidth}
  \centering
      \includegraphics[width=\linewidth]{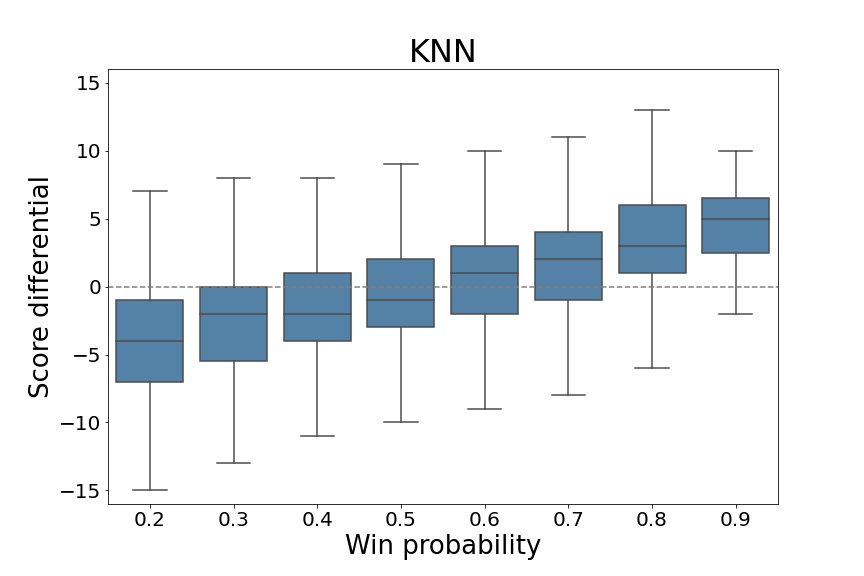}
\end{subfigure}
\begin{subfigure}{.5\textwidth}
  \centering
      \includegraphics[width=\linewidth]{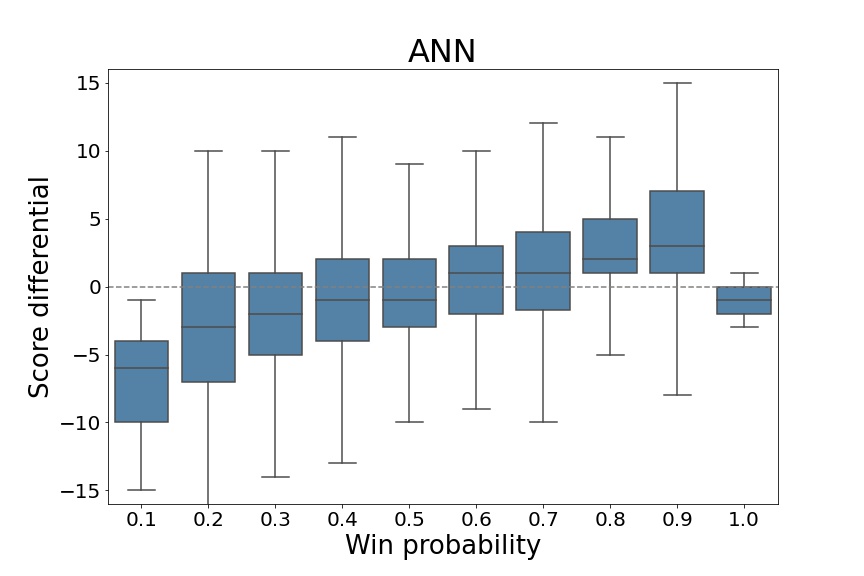}
\end{subfigure}%
\begin{subfigure}{.5\textwidth}
  \centering
      \includegraphics[width=\linewidth]{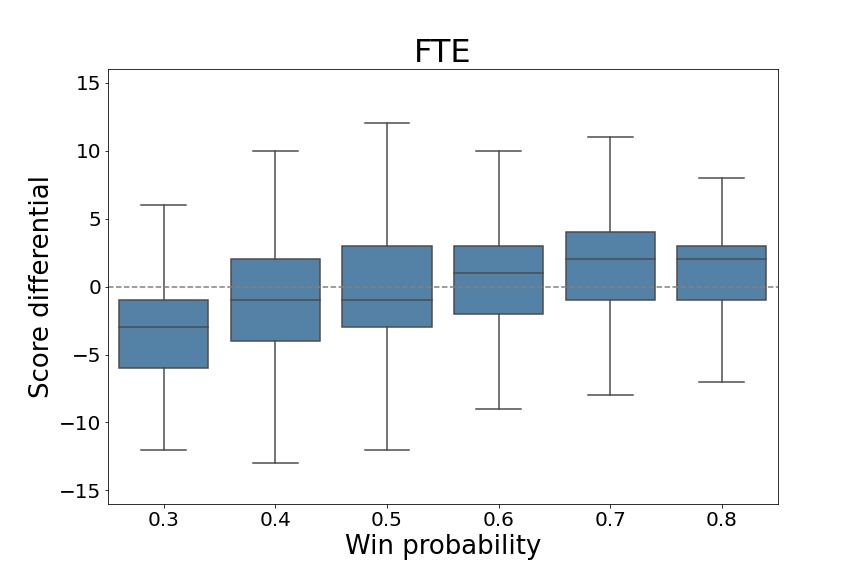}
\end{subfigure}
\caption{Predicted home team win probability versus score differential (home team final score minus away team final score). All predicted win probabilities are unscaled and rounded to the nearest 10\%. A dashed horizontal line at zero score differential (tied game) is shown for reference. LogR, KNN and ANN show the strongest positive linear trends. XGB and KNN perform the best for games with the most extreme predicted win probabilities. Note that SVM, KNN and FTE all fail to predict any games to have an especially high or low win probability.}
\label{fig:figure_2}
\end{figure}

\newpage

\begin{figure}[!h]
\includegraphics[width=\linewidth]{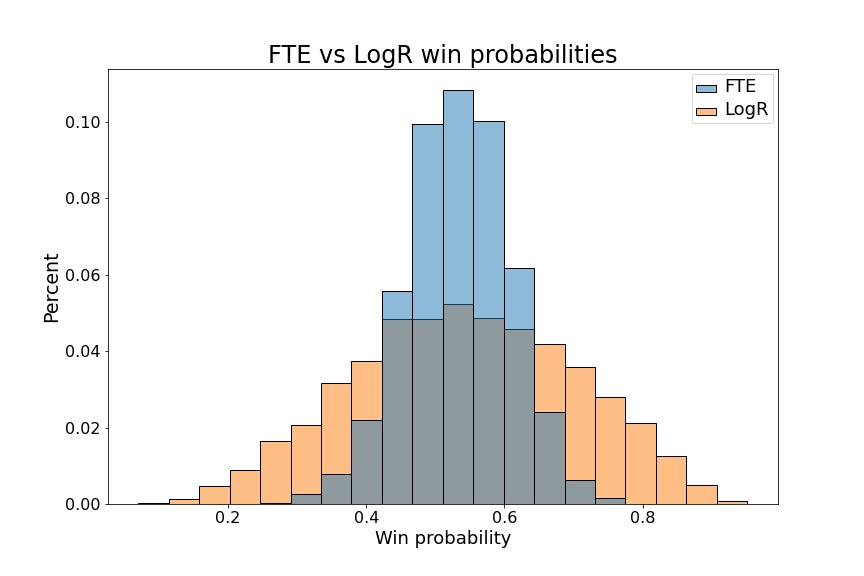}
\caption{Distribution of predicted home team win probabilities for FTE versus LogR. Note the narrow prediction region for FTE and the skinny tails. This means that many games will marked as being toss-ups.}
\label{fig:figure_3}
\end{figure}

\newpage

\begin{figure}[!h]
\centering
\includegraphics[width=\linewidth]{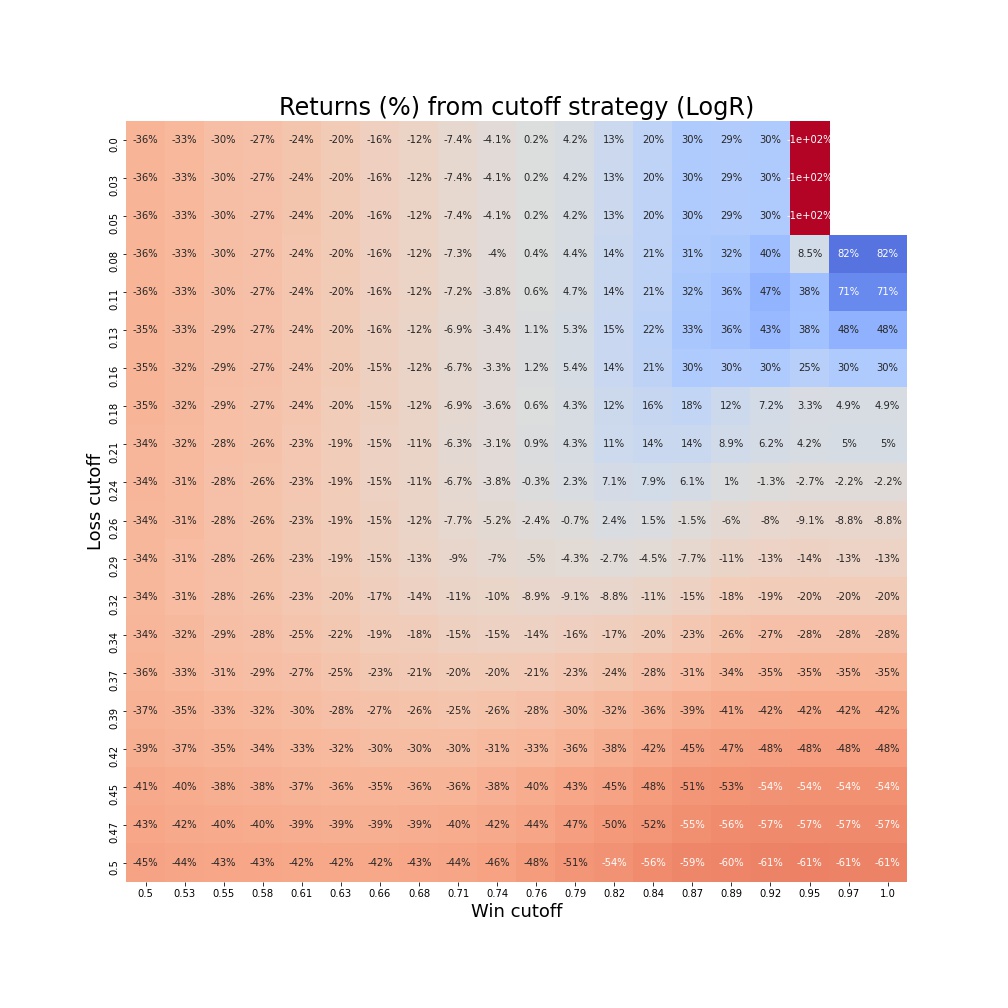}
\caption{Returns as a percentage of money invested using the high and low cutoffs shown on each axis. Predictions come from the LogR model. Setting the low and high cutoff equal to 0.5 is precisely the naive strategy described above. For appropriate choices of win and loss cutoffs we demonstrate positive returns, even into the double digits.}
\label{fig:figure_4}
\end{figure}

\newpage

\begin{figure}[!h]
\centering
\includegraphics[width=\linewidth]{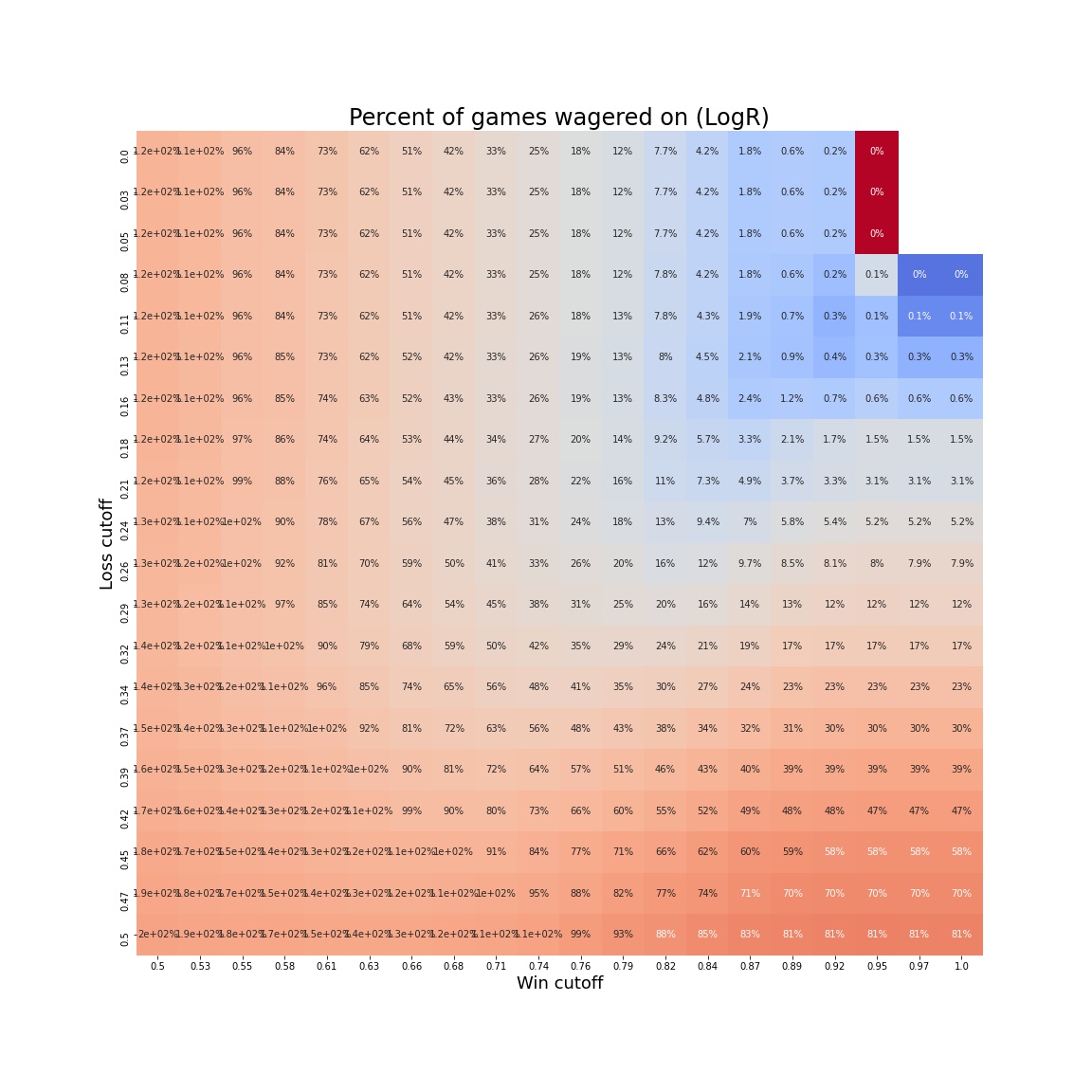}
\caption{Percentage of games wagered on using each pair of cutoffs using the LogR model. Colors correspond to the returns shown in the previous figure. Note that in situations where positive returns are realized, approximately 0.5\% to 5\% of games are wagered on. In a normal season of 2430 games this corresponds to between 5 and 120 games per season.}
\label{fig:figure_5}
\end{figure}

\end{document}